\documentclass[10pt,twocolumn,letterpaper]{article}

\usepackage{cvpr}
\usepackage{times}
\usepackage{epsfig}
\usepackage{graphicx}
\usepackage{amsmath}
\usepackage{amssymb}
\usepackage{subfigure}

\usepackage[pagebackref=true,breaklinks=true,letterpaper=true,colorlinks,bookmarks=false]{hyperref}

\cvprfinalcopy 

\def\cvprPaperID{593} 

\begin{document}

\setlength{\textfloatsep}{8pt plus 1.0pt minus 2.0pt}
\setlength{\abovecaptionskip}{1mm}
\setlength{\belowcaptionskip}{1mm}
\setlength{\abovedisplayskip}{1pt plus 2pt minus 6pt}
\title{Towards End-to-End Face Recognition through Alignment Learning}

\author{Yuanyi Zhong \qquad Jiansheng Chen \qquad Bo Huang \\
Department of Electronic Engineering, Tsinghua University\\
Beijing, China, 100084\\
{\tt\small zhongyy13@mails.tsinghua.edu.cn, jschenthu@mail.tsinghua.edu.cn, huangb14@mails.tsinghua.edu.cn}
}

\maketitle

\begin{abstract}

Plenty of effective methods have been proposed for face recognition during the past decade. Although these methods differ essentially in many aspects, a common practice of them is to specifically align the facial area based on the prior knowledge of human face structure before feature extraction. In most systems, the face alignment module is implemented independently. This has actually caused difficulties in the designing and training of end-to-end face recognition models. In this paper we study the possibility of alignment learning in end-to-end face recognition, in which neither prior knowledge on facial landmarks nor artificially defined geometric transformations are required. Specifically, spatial transformer layers are inserted in front of the feature extraction layers in a Convolutional Neural Network (CNN) for face recognition.  Only human identity clues are used for driving the neural network to automatically learn the most suitable geometric transformation and the most appropriate facial area for the recognition task. To ensure reproducibility, our model is trained purely on the publicly available CASIA-WebFace dataset, and is tested on the Labeled Face in the Wild (LFW) dataset. We have achieved a verification accuracy of 99.08\% which is comparable to state-of-the-art single model based methods. 

\end{abstract}

\section{Introduction}

During the past several years, the introduction of Convolutional Neural Networks (CNNs) have dramatically improved the state-of-the-art performances of many computer vision tasks including face identification and verification~\cite{sun2013hybrid,sun2014deep,sun2015deepid3,taigman2014deepface,schroff2015facenet,parkhi2015deep,wen2016discriminative}.
Instead of constructing classification models over traditional hand-crafted features, the data-driven nature of deep learning has successfully enhanced the robustness of learned facial features. 
As such, a well trained CNN is capable of handling pose, occlusion and illumination variations of face images to a considerably high degree \cite{schroff2015facenet,parkhi2015deep,ding2016comprehensive}.

\begin{figure}[b]
\begin{center}
\includegraphics[width=0.95\linewidth]{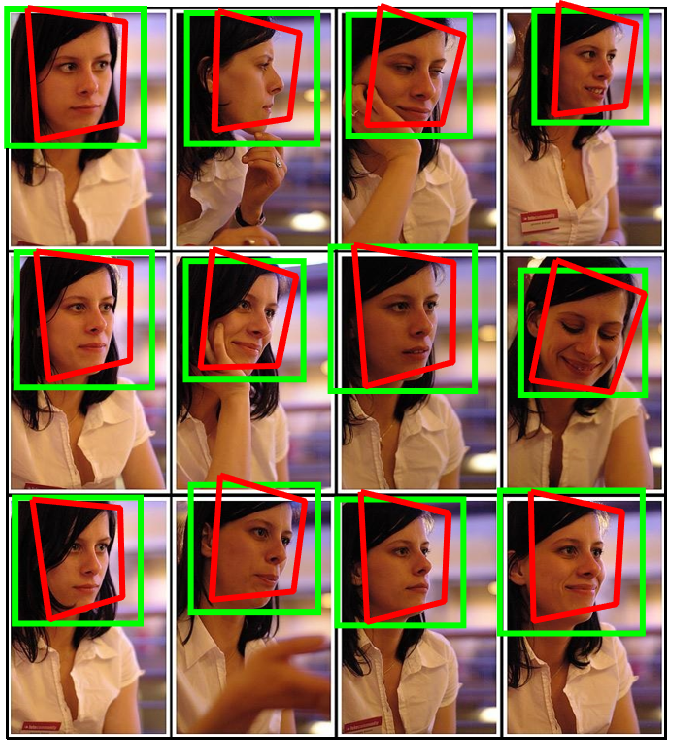}
\end{center}
   \caption{Sample alignment results on an image from the AFW~\cite{Zhu2012Face} dataset. The green rectangles are face detection results and the red rectangles show the model predicted projective transformations for the alignment of the faces .}
\label{fig01}
\end{figure}

However, large facial pose variation in real life scenarios still remains a challenge for practical face recognition systems. 
There are in general two kinds of intrinsically relevant approaches to handle this problem.
One approach is to build pose-aware or part-based models to handle face images of specific poses~\cite{masi2016pose,liao2013partial}. 
Another more commonly used approach is to introduce a explicit face alignment procedure before feature extraction for face recognition~\cite{sun2014deep,hu2016face,taigman2014deepface,ding2016comprehensive}.  
Previous studies~\cite{schroff2015facenet,parkhi2015deep} have already confirmed that performing explicit face alignment, especially in the testing stage, can efficiently improve the recognition performance. 
Therefore, a typical face recognition process usually consists of four stages: (1) detecting faces from captured images, (2) locating facial landmarks and aligning the detected faces using 2D or 3D geometric transformations, (3) extracting facial features for recognition, (4) deciding personal identity based on the likelihood between feature templates. 
Recent research shows that while 3D alignment may seem superior over 2D alignment, it shows no significant advantage in terms of recognition accuracy especially when CNNs are used to extract facial features~\cite{banerjee2016frontalize}.

There are two major concerns regarding such a framework in which face alignment and facial feature extraction are performed independently. 
First, most face alignment methods~\cite{xiong2013supervised,ren2014face,chen2014joint,zhang2016joint} rely on the accurate facial landmark location, a vision problem which may be even more difficult than face recognition considering that manual labeling of facial landmarks is much more laborious and expensive than collecting personal identity information. 
It is true that facial landmarks can be used in other interesting applications such as face synthesis and action unit analysis. 
What we argue here is that using the face landmark location as a prerequisite for the face recognition might not be necessary. 
What is more, under such a framework, the redefinition of landmarks and relabeling of training data are inevitable for other fine-grained classification tasks such as the animal recognition~\cite{zhang2013deformable,zhang2014part}.
Second, the principles of geometric transformation are usually artificially defined in the face alignment. 
For example, a widely used stratagem is to align the landmarks around eyes and mouths through non-reflective similarity transformation. 
However, it is not clear whether the succeeding facial feature extraction can benefit from other kinds of 2D transformation based on other landmarks. 

Therefore, the essential question is that since that facial features for recognition can be successfully learnt in a data-driven manner, why can't the face alignment be? 
After all, it seems quite weird to still rely on artificial priors in the alignment process while all the other stages of face recognition have been proved to be data-driven trainable. 
Hence, in this paper, we try to propose a deep learning face recognition model in which a Spatial Transformer module \cite{jaderberg2015spatial} is used to supersede the process of face alignment, so that the face alignment and recognition can be unified to the same homogeneous framework. 
The proposed model is end-to-end trainable and it does not require any explicit knowledge about the characteristics of the human face nor artificially defined alignment principles.
During training, the model automatically learns a consistent way of aligning each face image so as to best suit face recognition purposes based on identity clues only. 

We observed through experiments that the proposed model generally tends to transform a human face to a upright standard position, just as people have been doing heuristically in existing methods~\cite{sun2014deep,yi2014learning}. 
This is not surprising considering that most faces in real-life images are nearly upright. 
Figure~\ref{fig01} shows the projective transformations predicted by our model for faces with large pose variations. Intuitively, the model predictions comply well with the underlying ground truth transformations.
This is interesting considering that no supervision signal on the ground truth transformations was ever used during model training.

There are several advantages of the proposed method. 
Through an end-to-end learning, the face alignment and the facial feature extraction may interact with each other so as to achieve a joint optimum in term of the recognition task. 
The adaptation ability to environment and capturing device changes of face recognition can also be enhanced. 
What is more, the learnt transformation and intermediate facial images can be readily used for other purposes. 
For example, the upright normalized face images might facilitate more accurate facial attribute prediction and landmark localization. 
Moreover, the model can be easily extended to other fine-grained image classification problems. 
The main contributions of this work can be summarized as follows.
\begin{enumerate}
	\item We propose a face recognition system which requires neither independent face alignment process nor prior knowledge about facial structures.
	\item We demonstrate that the proposed face recognition model is end-to-end trainable through standard SGD.
	\item We reveal that no prior knowledge about human faces is necessary for face recognition! This indicates that despite the neuroscience evidences for functional organization in human face perception~\cite{Kanwisher1997The}, whether and how face perception differs from generic fine-grained object perception remains an open question.
\end{enumerate}

The paper is organized as follows. In section 2, we briefly present several closely related existing works. In section 3, we describe the details of our model architecture. Experimental results on the LFW~\cite{LFWTech} dataset and the YTF~\cite{Wolf2011Face} dataset are presented in section 4. Section 5 concludes our work.


\section{Related Work}

Recently, the introduction of deep learning models has greatly promoted the development of the face recognition technology. 
Records of recognition accuracies on several major challenging benchmarks have been constantly broken since the Facebook DeepFace system~\cite{taigman2014deepface} demonstrated the effectiveness of the data driven deep learning paradigm for face recognition. 
A lot of Deep models of various structures, especially CNNs, have been proposed for face recognition ever since then. 
Exemplary works include the DeepID~\cite{sun2014deep,sun2015deeply,sun2015deepid3} track of models, the FaceNet~\cite{schroff2015facenet}, etc. 
It has been widely accepted that comparing to traditional handcrafted features such as the high dimensional LBP~\cite{chen2013blessing} or features learnt by imposing artificially designed constraints such as the Bayesian face~\cite{chen2012bayesian} and the GaussianFace~\cite{lu2014surpassing}, automatically learnt deep features 	 based directly on personal identity clues are obviously more advantageous in terms of both the discriminative power and robustness. 

In most deep learning based face recognition methods, the inputs to the deep model are aligned face images during both training and testing. 
Typically, the alignment is performed by fitting a 2D or 3D~\cite{hu2016face,taigman2014deepface} geometric transformation between the positions of detected facial landmarks and certain predefined landmarks. 
It has been revealed that a proper alignment is crucial to the recognition performance. 
Parkhi et. al. reported a 1\% recognition accuracy improvement on the LFW dataset when alignment was adopted in the testing stage~\cite{parkhi2015deep}. 
It has also been shown that in terms of the recognition accuracy, 3D frontalization does not show significant improvements over simple 2D alignments~\cite{banerjee2016frontalize}. 
Therefore, we will focus on the 2D image alignment only in this paper.

In the work of the FaceNet~\cite{schroff2015facenet}, Sharoff et. al. tried to avoid the explicit face alignment and used an extremely large training dataset consisting of over 200 million images of 8 million identities to realize highly pose-invariant facial feature extraction. 
Nevertheless, they still found that adding additional face alignments during the testing stage may further increase the recognition accuracy. 
To a certain extents, this shows the indispensability of the explicit face alignment even in a deep learning based face recognition framework. 
Interestingly, despite of its significance, the face alignment is usually implemented based on an independent landmark location process and several artificially defined transformation principles in most existing systems. 
The automatic learning of optimum geometric transformations for face recognition has been largely overlooked although the  face alignment process has already become the greatest impediment towards an end-to-end training of face recognition models.

Actually, the problem of learning geometric transformations has already been studies in other computer vision tasks such as the handwritten digits recognition and the bird classification~\cite{jaderberg2015spatial}.  
More specifically, Jaderberg et. al. introduced a differentiable CNN component call the Spatial Transformer, which aims at improving the robustness of CNN towards translation, scaling, rotation and even more generic image warping structures~\cite{jaderberg2015spatial}. 
Due to its differentiability, a spatial transformer can be trained to learn the optimum parametrized transformation for a particular computer vision task based on a specific feature map through backward propagation. 
Very recently, Chen et. al. successfully used the spatial transformer in a supervised manner for boosting the performance of the face detection~\cite{chen2016supervised}. 
Inspired by this work, we propose to use spatial transformers for enabling the simultaneous learning of the optimum face alignment together with the facial feature extraction for the face recognition. 
A previous work seemingly similar to our proposal is~\cite{tadmor2016learning}, in which a neural network was used to predict the transformation parameters so as to facilitate face recognition on an embedded platform. 
However, this neural network is trained in a supervised manner using artificially defined transformation parameters as the ground truth. 
While our target is to enable the automatic learning of the optimum geometric transformation for face recognition driven by personal identity clues only.
As such, the artificial definition of the transformation form will no longer be needed and the end-to-end training of face recognition models can be facilitated.


\section{Methodology}

This section presents the overall design of the proposed end-to-end face recognition system. 
The general architecture of our system is described first. In particular, the design scheme of the localization network which is used for predicting geometric transformation parameters is discussed. 
To ensure the completeness and reproducibility of this paper, we then elaborate the details of the spatial transformer layer~\cite{jaderberg2015spatial} for different transformation types. 
The effect of the transformation type selection on the face recognition performance is also inspected.

\subsection{System Architecture}
		
\begin{figure*}[htbp]
\begin{center}
\includegraphics[width=0.95\linewidth]{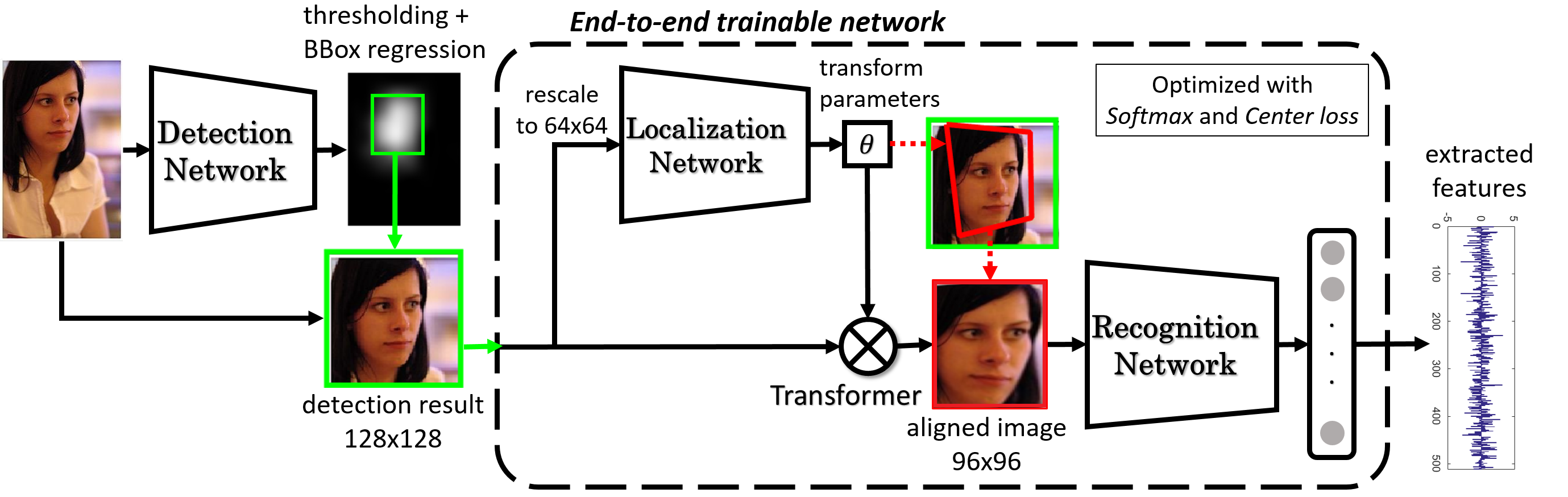}
\end{center}
\caption{The overall architecture  of the proposed face recognition system.}
\label{fig02}
\end{figure*}
	
Generally speaking, a typical face recognition system takes camera captured images or video sequences as the input and deliver located faces along with their identities as the output.
Nowadays, a face recognition system is commonly divided into three major components of detection, alignment and recognition.
These three components are usually designed and trained separately. 
A possible historical reason for this situation is that under the traditional technology framework, different mathematic models of heterogeneous structures are found to be suitable for these three seemingly different computer vision tasks. 
The concatenation and unified training of these models can be prohibitively complex or even intractable.

However, recent research results have already confirmed the effectiveness of CNNs on face detection~\cite{zhang2016joint}, landmark location~\cite{Zhou2013Extensive} and recognition~\cite{sun2014deep,taigman2014deepface}.
More interestingly, the network architecture of the CNNs used in these different tasks can be as  similar to each other as necessary.
This has actually made the design and implementation of an end-to-end face recognition model technically possible.
Ideally, such a model should be trained in an fully end-to-end manner using only identity clues in the input images as supervising information.
As such, the optimum image regions as well as the optimum transformation imposed on them can be simultaneously learned to benefit personal identity recognition. 
Training of such a model can also be extremely difficult.
In order to simplify the problem, we keep the face detection as an independent task and focus on the end-to-end design and implementation of the alignment and recognition only, as is shown in Figure~\ref{fig02}. 
Such a design is consistent with the hypothesis in cognitive neuroscience that face detection and identification might use separate dedicated resources and mechanisms in human brains~\cite{Tsao2008Mechanisms}.


For the face detection task, we stacked two additional layers for the face saliency map prediction and the face bounding box regression after the second inception module of the GoogLeNet~\cite{Szegedy2015Going}, and fine-tuned the network with weights initialized from the original GoogLeNet model on the publicly available WIDER dataset~\cite{yang2016wider}. 
Similar approach has be adopted in the UnitBox method~\cite{Yu2016UnitBox}, where the fine-tuned VGG net was used instead of the GoogLeNet. 
The performance of this simple and straight-forward approach is quite satisfactory. 
A recall rate of 83\% at 200 false alarms has been achieved on the FDDB~\cite{fddbTech} dataset.
Sample face detection results are also shown in Figure~\ref{fig01}.
Actually, any off-the-shelf face detector such as the MTCNN~\cite{zhang2016joint} can be probably used in our system. 
We have found through experiments that the effect of the face detection accuracy on the final recognition performance is surprisingly low. 
This is largely because that the localization network can successfully learn a proper way for tolerating the possible inaccuracy or instability in the detection bounding box positions.

For the alignment and recognition task, we design an end-to-end network consisting of mainly three parts: a localization network which predicts the 2D transformation parameters based on the down-sampled input face image; a sampler which warps the face image according to the predicted transformation parameters and a deep facial feature extraction network for recognition. The data flow and intermediate results of this network are shown in Figure~\ref{fig02}. 

During the training stage, the detected face bounding boxes and the personal identity information are used for supervision.
More specifically, the face regions are first cropped according to the detected bounding boxes.
Then the input to the network are these face crops after being resized to $128\times 128$ pixels.
For the localization network, we adopt a neural network with 3 convolutional layers with kernel sizes of $5\times5,3\times3$ and $3\times3$ respectively. The PReLU and $2\times2$ pooling are used after each convolution layer.
After that, one fully connected layer of the size of 64 is used before the regression of the geometric transformation parameters (8 parameters for projective; 6 for affine; 4 for similarity).
The input face crops are down-sampled to $64\times64$ pixels before being fed into the localization network since we have observed that high resolution image is generally not necessary for computing the transformation parameters. 
Inspired by~\cite{wen2016discriminative}, we use the similar deep residual network (ResNet)~\cite{he2015deep} for recognition feature extraction and representation learning considering the high generalization capability of the ResNet demonstrated in various visual recognition problems. 
The residual network consists of 9 residual blocks with 24 convolution layers in total, producing a 512 dimensional output feature vector which is presumably able to capture the intra-personal variations holistically. 
The CenterLoss function proposed in~\cite{wen2016discriminative} is also used along with the SoftMax loss for learning discriminative features for the recognition.

\subsection{Localization Network}

The designing of the localization network is essentially a tradeoff between the structural complexity and the prediction accuracy. 
In order to facilitate the end-to-end training, simple structures are favorable so long as sufficient prediction accuracies can be ensured.
Therefore, we conducted an experiment to decide the best localization network architecture.
We first computed the affine transformation parameters by fitting facial landmarks to a set of predefined locations in the traditional way on CASIA WebFace~\cite{yi2014learning} images and LFW images.
Then we used the WebFace images to train a series of transformation networks with increasing structural complexity while keeping the total number of parameters to be roughly the same. 
We then tested the generalization ability of the trained networks on the LFW images.
The network architecture with small enough fitting errors and modest complexity is adopted. 
Details of the network designs as well as the fitting errors are shown in Table~\ref{tableLoc}.
According to the results, we chose the network consisting of 3 convolutional layers and 1 fully connected layers, each with a PReLU layer inserted afterwards, in our implementation.

	\begin{table} 
		\caption{Comparison of localization network architectures}
		\label{tableLoc}
		\centering
		\begin{tabular}{ccc}
			\hline
			Architecture & \#Params & Fitting Error (MSE)\\
			\hline
			1 Conv+Pool, 1 FC & 340K & 0.004980 \\
			2 Conv+Pool, 1 FC & 313K & 0.004689 \\
			\textbf{3 Conv+Pool, 1 FC} & \textbf{277K} & \textbf{0.004587} \\
			4 Conv+Pool, 1 FC & 260K & 0.005188 \\
			2 Conv+Pool, 2 FC & 314K & 0.004852 \\
			3 Conv+Pool, 2 FC & 280K & 0.004747 \\
			\hline
		\end{tabular}
	\end{table}

\subsection{Spatial Transformer Network}

According to the original DeepMind paper~\cite{jaderberg2015spatial}, the spatial transformer can be used to implement any parametrizable transformation including translation, scaling, affine, projective, and even thin plate spline transformation.
In traditional face alignment implementations, the similarity transformation is most commonly adopted. 
However, Wagner et. al. showed the effectiveness of the projective transformation for handling large pose variation in the robust face recognition~\cite{Wagner2012Toward}. 
To ensure completeness, we investigate three types of homogeneous transformations, namely similarity, affine and projective, in this work.
Considering the fact that Jaderberg et. al. only elaborated the detailed implementation of the affine transformation in~\cite{jaderberg2015spatial}, we hereby briefly review the structure of the Spatial Transformer, and take projective transformation and similarity transformation as examples to demonstrate their forward and backward computations.

Suppose that for the $i_{th}$ target point $\mathbf{p^t_i} = (x^t_i,y^t_i,1)$ in the output image, a grid generator generates its source coordinates $(x^s_i,y^s_i,1)$ in the input image according to transformation parameters. 
For the projective transformation, such a process can be expressed by (\ref{eq_projective}) in which $A$ to $H$ are eight transformation parameters and $z^s_i = G x^t_i + H y^t_i + 1$.

\begin{equation}\label{eq_projective}
\left(\begin{matrix}
	x^s_i\\y^s_i\\1
	\end{matrix} \right) = 
	\mathcal{T}\circ \mathbf{p^t_i} = 
	\frac{1}{z^s_i}
	\left( \begin{matrix}
	A&B&C\\D&E&F\\G&H&1
	\end{matrix} \right)
	\left(\begin{matrix}
	x^t_i\\y^t_i\\1
	\end{matrix} \right)
\end{equation}
	
Then the sampler samples the input image $U$ at generated source coordinates.
This is equivalent to convolving a sampling kernel $k$ with the source image of height $H$ and width $W$ as is shown in (\ref{eq_sample}) in which $V_i$ stands for the pixel value of the $i_{th}$ point in the output image.
We use the bilinear kernel so that $k(w-x_i^s,h-y_i^s)=\max(0,1-|w-x_i^s|) \times \max(0,1-|h-y_i^s|)$.
 
\begin{equation}\label{eq_sample}
V_i = \sum_{h=1}^{H} \sum_{w=1}^{W} U_{wh} k(w-x_i^s,h-y_i^s) 
\end{equation}

During the backward propagation, we need to calculate the gradient of $V_i$ with respect to each of the eight transformation parameters. The function shown in (\ref{eq_sample}) may not be differentiable when $w=x_i^s$ or $y=y_i^s$. However, this seldom happens since the chance that the calculated $x_i^s$ or $y_i^s$ are integers is very low in practice. We empirically set the gradient at these points to be $0$ considering that their effect on the back propagation process are negligible. For differentiable points, the chain rule can be applied to get the gradient. An example regarding $G$ is shown in (\ref{eq_devG}), in which the gradient with respect to the source coordinates are defined in (\ref{eq_devsrc}) and (\ref{eq_devsgn}). 

	\begin{equation}\label{eq_devG}
	\begin{split}
	\frac{\partial V_i}{\partial G} &= \frac{\partial V_i}{\partial z_i^s} \frac{\partial z_i^s}{\partial G} 
	= \left(\frac{\partial V_i}{\partial x_i^s} \frac{\partial x_i^s}{\partial z_i^s} + \frac{\partial V_i}{\partial y_i^s} \frac{\partial y_i^s}{\partial z_i^s}\right) x_i^t \\
	&=-\frac{x_i^t}{z_i^s}\left(\frac{\partial V_i}{\partial x_i^s} x_i^s + \frac{\partial V_i}{\partial y_i^s} y_i^s\right)
	\end{split}
	\end{equation}
	
	\begin{equation}\label{eq_devsrc}
	\begin{split}
	\frac{\partial V_i}{\partial x_i^s}
	&= \sum_{h=1}^{H} \sum_{w=1}^{W} U_{wh} \frac{\partial}{\partial x_i^s}k(w-x_i^s,h-y_i^s) \\
	&= \sum_{h=1}^{H} \sum_{w=1}^{W} U_{wh} \max(0,1-|h-y_i^s|) sg(w-x_i^s)
	\end{split}
	\end{equation}
	
	\begin{equation}\label{eq_devsgn}
	sg(w-x_i^s) = 	\left\{                 
	\begin{array}{lll}  
		0, \ |w-x_i^s|>1 \\ 
		1, \ 0 \leq w-x_i^s \leq 1 \\
		-1, \ -1 \leq w-x_i^s \leq 0
	\end{array}       
	\right. 
	\end{equation}

	The similarity transformation is defined in (\ref{eq_similarity}) in which $\alpha$ is the rotation angle, $\lambda$ is the scaling factor, and $t_1,t_2$ are the horizontal and vertical translation displacements respectively. Analogously, the gradients of $V_i$ respected to $\alpha$ and $\lambda$ are shown in (\ref{eq_devalpha}) and (\ref{eq_devlambda}) respectively.
	
	\begin{equation}\label{eq_similarity}
	\left(\begin{matrix}
	x^s_i\\y^s_i\\1
	\end{matrix} \right) = 
	\mathcal{T}\circ \mathbf{p^t} = 
	\left( \begin{matrix}
	\lambda \cos\alpha & -\lambda \sin\alpha & t_1\\
	\lambda \sin\alpha & \lambda \cos\alpha & t_2
	\end{matrix} \right)
	\left(\begin{matrix}
	x^t_i\\y^t_i\\1
	\end{matrix} \right)
	\end{equation}

	\begin{equation}\label{eq_devalpha}
	\frac{\partial V_i}{\partial \alpha} = \frac{\partial V_i}{\partial x_i^s} \frac{\partial x_i^s}{\partial \alpha} + \frac{\partial V_i}{\partial y_i^s} \frac{\partial y_i^s}{\partial \alpha}
	= \frac{\partial V_i}{\partial x_i^s} (t_2-y_i^s) + \frac{\partial V_i}{\partial y_i^s} (x_i^s-t_1)
	\end{equation}
	
	\begin{equation}\label{eq_devlambda}
	\begin{split}
	\frac{\partial V_i}{\partial \lambda} &= \frac{\partial V_i}{\partial x_i^s} \frac{\partial x_i^s}{\partial \lambda} + \frac{\partial V_i}{\partial y_i^s} \frac{\partial y_i^s}{\partial \lambda} \\
	&= \frac{\partial V_i}{\partial x_i^s} (x_i^t \cos\alpha - y_i^t \sin\alpha) + \frac{\partial V_i}{\partial x_i^s} (x_i^t \sin\alpha + y_i^t \cos\alpha)
	\end{split}
	\end{equation}

	

As mentioned in the introduction section, the most widely used alignment scheme for face recognition is the non-reflective similarity transformation. 
Nonetheless, it is by far unclear how different kinds of 2D transformations may affect the face recognition performance.
In order to explore the most suitable transformation type for face recognition, we will train four models with four different kinds of transformations namely identical, similarity, affine and projective, while keeping the training set and the rest of the network architecture unchanged. 
For the identical transformation, the detected facial region is directly cropped in the center for recognition and no substantial transformation is made.
The corresponding results and face verification accuracies on LFW and YTF will be shown in Section 4.1.
	
\subsection{Discussions}

In fact, it is possible to implement a fully end-to-end face recognition system based on our proposed framework.
The face detection stage can actually function as a Region Proposal Network~\cite{Ren2016Faster} or an attention model~\cite{Xiao2015The} to propose candidate facial regions, so that it can be easily connected to the proposed alignment and recognition network described above.
Also, although the spatial transformers are differentiable, the gradient descend methods may not be the only way for training them. The reinforcement learning~\cite{Caicedo2015Active} based approach can potentially be applied to train the network in a more efficient way.

\section{Experimental Results}

Two sets of experiments are described in this section for demonstrating the effectiveness of the proposed method. 
First, recognition experiments are performed on the LFW and YTF datasets.
Previous works have already confirmed that the face recognition accuracy can be effectively enhanced either by increasing the training set size~\cite{schroff2015facenet} or by fusion of multiple deep models in an ensemble way~\cite{sun2015deepid3}. 
However, in this work we mainly focus on studying the feasibility of the proposed end-to-end architecture as well as how different transformation types in face alignment may affect the recognition task. 
Therefore, we only use the CASIA-Webface images for training the alignment and recognition models with different transformation types and only use a single deep model in recognition.
This somehow also ensures the reproducibility of this work.  
Second, we test the effectiveness of using our model predicted transformations for improving the facial landmark location accuracy of existing algorithms.

\subsection{Recognition Experiments}

	\begin{table}[t]
		\setlength{\tabcolsep}{3pt}
		\caption{Face verification performance on LFW and YTF datasets}
		\label{tablePerform}
		\centering
		\begin{tabular}{lcccc}
			\hline
			Method & Trainset & \#Models & LFW & YTF\\
			\hline
			DeepFace~\cite{taigman2014deepface} & 4M & 3 & 97.35\% & 91.4\% \\
			FaceNet~\cite{schroff2015facenet} & 200M & 1 & 99.63\% & 95.1\% \\
            DeepID~\cite{sun2014deep2} & 0.2M & 100 & 97.45\% & - \\
            DeepID2+~\cite{sun2015deeply} & 0.2M & 25 & 99.47\% & 93.2\% \\
            VGG Face~\cite{parkhi2015deep} & 2.6M & 1* & 99.13\% & 97.3\% \\
            Center face~\cite{wen2016discriminative} & 0.7M & 1 & 99.28\% & 94.9\% \\
            \hline
            ResNet (Pre-aligned) & 0.46M & 1 & 98.35\% & - \\
	    Ours (Identical) & 0.46M & 1 & 97.68\% & 92.9\% \\
	    Ours (Similarity) & 0.46M & 1 & 98.65\% & 94.6\% \\
	    Ours (Affine) & 0.46M & 1 & 98.71\% & 94.7\% \\
            Ours (Projective) & 0.46M & 1 & 99.08\% & 94.7\% \\
			\hline
		\end{tabular}
		* the facial feature is the average of 30 multiscale patches
	\end{table}

\begin{figure}[b]
\begin{center}
\includegraphics[width=0.95\linewidth]{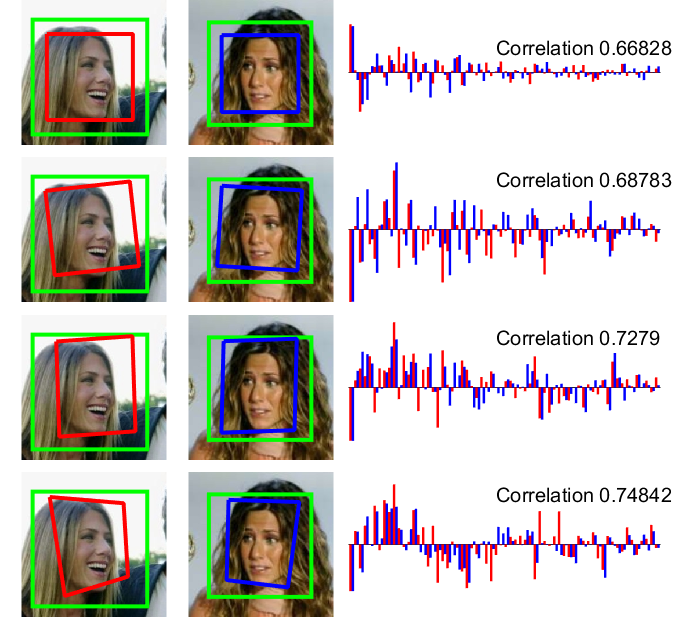}
\end{center}
   \caption{Comparison of predicted transformations and extracted facial features for a pair of Jennifer Aniston's images from LFW using models of different transformation types. For the top row to the bottom row are identical, similarity, affine and projective respectively. The green rectangles are face detection results and the red/blue quadrangles shows the predicted transformations. On the right side shows the first 64 principle components of the extracted feature vectors along and their corresponding correlation coefficients (cosine similarities).}
\label{fig03}
\end{figure}

We trained our proposed end-to-end network on a cleaned version of the CASIA-WebFace dataset, which consists of around 10k unique identities and 460k face images. 
The horizontally flipped images were used for data augmentation during training.
For each WebFace image, the face detector described previously in Section 3 was used to locate the facial region. 
Then the original image was cropped using a slightly enlarged version of the detected bounding box. 
The cropped images were used as the training inputs to the end-to-end alignment and recognition network.
For face detection failures, we simply used the fixed size center crops of the original images.

We set the batch size to 100 images for each training iteration.
We jointly used the center loss and the softmax loss. 
The coefficient of the center loss relative to the softmax loss was set to 0.008, as recommended by~\cite{wen2016discriminative}.
The learning rate of the recognition network is set to 0.01 and decays after every 10000 iterations.
We observed during experiments that the best training results were achieved when the learning rate of the localization network was 10 to 100 times smaller than that of the recognition network.
This can be understood considering that the loss value of the recognition network is nearly one to two magnitudes larger than the values of the transformation parameters in practice.
The training process took around 8 hours on a nVidia TitanX GPU after about 100,000 iterations.

\begin{figure}[b]
\begin{center}
\includegraphics[width=0.9\linewidth]{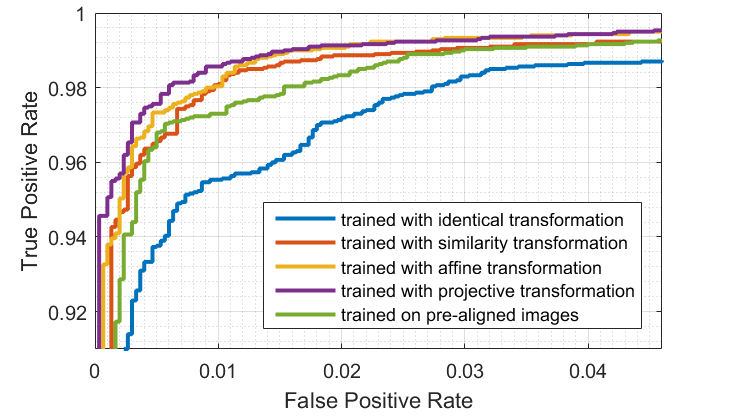}
\end{center}
   \caption{ROC curve for face verification on LFW}
\label{fig04}
\end{figure}

\begin{figure}[b]
\begin{center}
\includegraphics[width=0.9\linewidth]{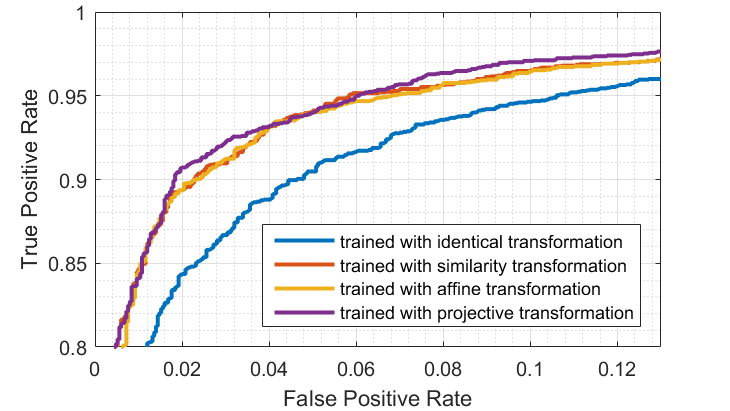}
\end{center}
   \caption{ROC curve for face verification on YTF}
\label{fig05}
\end{figure}

Although the proposed network is trained in an end-to-end manner,
we have found that it is helpful to randomly reinitialize the parameters of the recognition network once in the middle of the training process.	
This can be possibly caused by the huge disparity in the structural complexity between the localization network and the recognition network.
It is generally easy for the localization network to be stabilized near its global optimum at the early stage of the training due to its relative simple architecture. 
On the contrary, the chance that the recognition network falls into its local optimum before the localization network is stabilized is quite high considering its highly complex structure.
Therefore, reinitialization of the recognition network provides a chance to let it escape from local optima.

We tested our models in the face verification task of two widely used unconstrained face recognition benchmark datasets, namely LFW and YTF. LFW contains 13233 images from 5749 people, requiring verification of 6000 face pairs; and YTF consists of 3425 videos of 1595 people, requiring verification of 5000 video pairs.
Both datasets allow 10-fold cross validation according to the standard \emph{unrestricted with labeled outside data} protocol. 
We averaged the two feature vectors of each test image and its mirrored version as the deep feature representation. 
The similarity score between a pair of images was computed using the cosine distance between the corresponding feature vectors after dimension reduction using PCA.
We train the PCA and select the best classification threshold based on the 9 training folds, and then test on the remaining test fold.

Table~\ref{tablePerform} shows the numerical results of the verification performances.
For a fair comparison, we also independently trained the ResNet based recognition network on the pre-aligned (normalized) images provided by the WebFace dataset. 
Several observations can be made according to the verification accuracies. 
First, among the four types of transformations, the identical transformation results in the lowest verification accuracy (97.68\% \& 92.9\%).
This is consistent with the conclusion in previous works that a explicit alignment of face images may significantly benefit the face recognition.
Second, although sharing the same underlying recognition network structure, the model trained on the artificially defined alignment (98.35\%) is inferior to those trained on learned alignments in terms of verification accuracy.
This demonstrates the advantage of the proposed end-to-end joint training of alignment and recognition.
Third, comparing to the similarity transformation (98.65\%) and the affine transformation (98.71\%), the project transformation (99.08\%) should be more suitable for face recognition.
This is not surprising considering that the projective transformation can describe the camera imaging process of most face images more accurately.

Figure~\ref{fig03} visually illustrates the effect of different transformations on the face feature extraction.
Intuitively, a tendency can be observed that more complex transformation types usually lead to higher robustness of extracted face features especially for images with large pose variations.
Figures~\ref{fig04} and \ref{fig05} show the corresponding ROC curves.
It can be observed that comparing to the LFW dataset, the effect of different transformation types on the verification accuracy is much less significant on the YTF dataset.
The curves for the similarity and affine transforms almost overlap with each other.
This is expected since the face pose variation problem is largely alleviated by averaging the facial features extracted from a sequence of images in the video frames on YTF.
Even though, the explicit face alignment still helps.

\subsection{Landmark Location Experiments}

The transformations predicted by our model can be used to normalize the face region to a near frontal canonical view.
Besides recognition, the normalized face images may be used to improve the accuracy of other tasks such as the gender recognition, the expression classification and the landmark location.
In most existing face recognition systems, the landmark location usually serves as a foundation for the face alignment.
However, we will demonstrate here that the face alignment can be conversely used to improve the accuracy of the facial landmark location especially for the methods that are relatively susceptible to pose variations.

\begin{figure}[t]
\begin{center}
\includegraphics[width=0.9\linewidth]{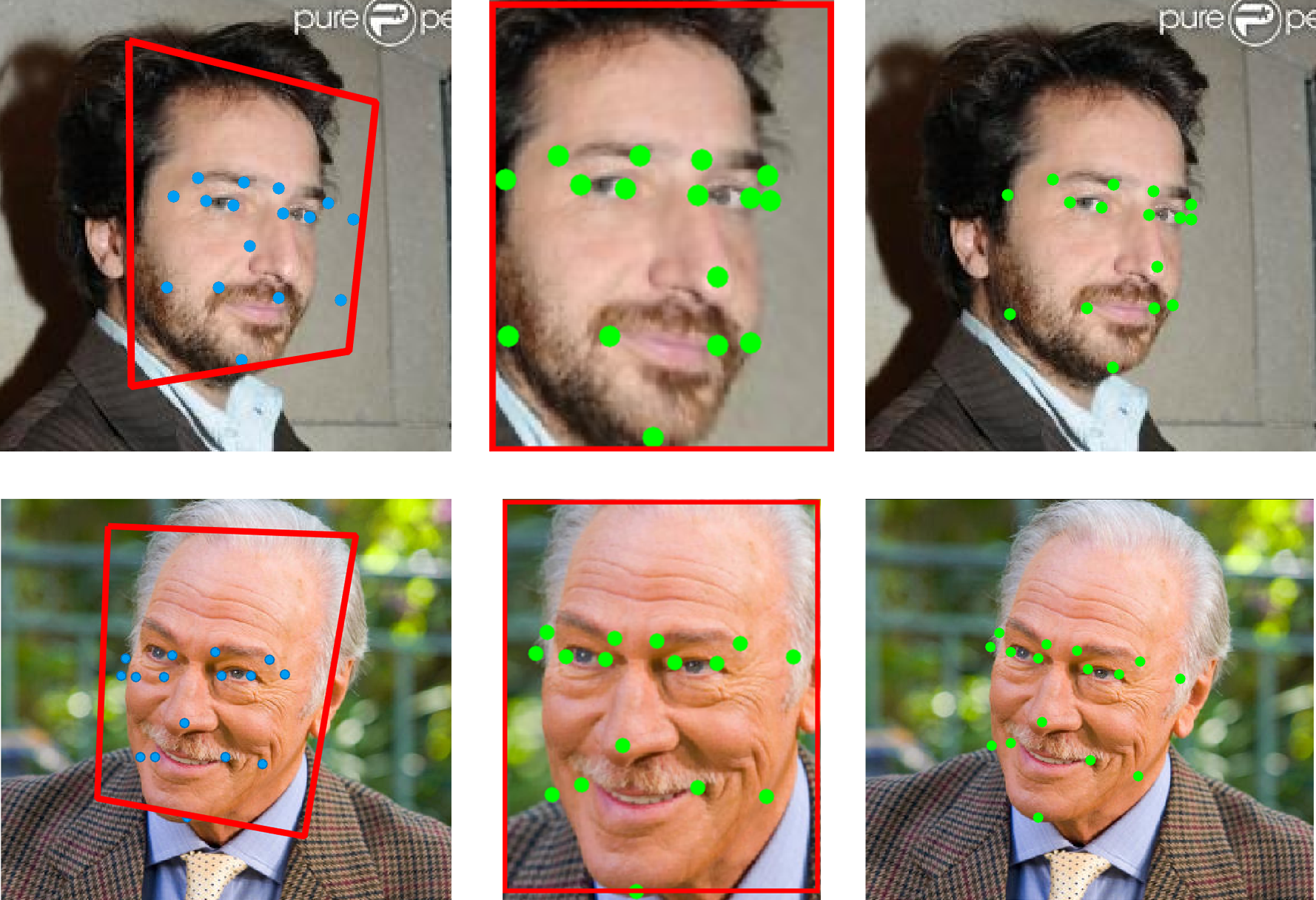}
\end{center}
   \caption{Improving landmark location using predicted transformations. 1st column: ASM results (cyan dots) on the original images and the predicted projective transformations (red quadrangles). 2nd column: normalized face images using predicted transformation and the relocated landmarks. 3rd column: relocated landmarks projected back to the original images.}
\label{fig06}
\end{figure}

\begin{figure}[b]
\begin{center}
\includegraphics[width=0.9\linewidth]{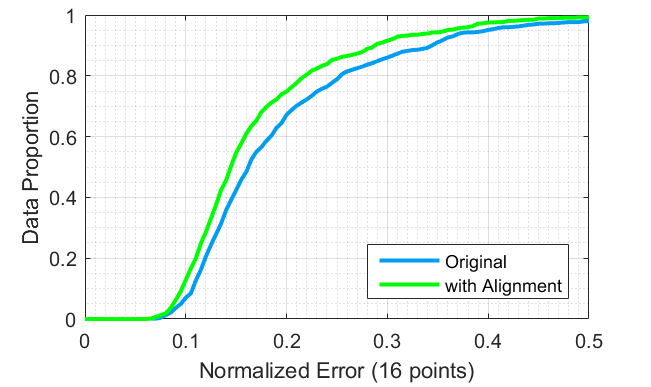}
\end{center}
   \caption{CED curves on LFPW dataset.}
\label{fig08}
\end{figure}

The basic idea is simple and is shown in Figure~\ref{fig06}. 
After the face alignment, the landmark location is performed on the normalized face image instead of the original image. 
Then the coordinates of the points are mapped back to the original image using the geometric transformation for alignment. 
We chose a typical implementation~\cite{Milborrow2014} of the Active Shape Model (ASM) based method as example. 
From Figure~\ref{fig06} we can see obvious inaccuracies of the original landmarks caused by facial pose variations, and significant improvements can be achieved with the help of the face alignments.
We tested the landmark location accuracy on the LFPW dataset~\cite{Belhumeur2011Localizing}.
The location error was measured as the average distance between the 16 labeled and located landmarks.
Such error is normalized by the inter-ocular distance.
Figure~\ref{fig08} compares the Cumulative Error Distribution (CED) curves with and without alignment.
Obvious improvement has been achieved by using alignment.

\begin{figure}[t]
\begin{center}
\includegraphics[width=0.9\linewidth]{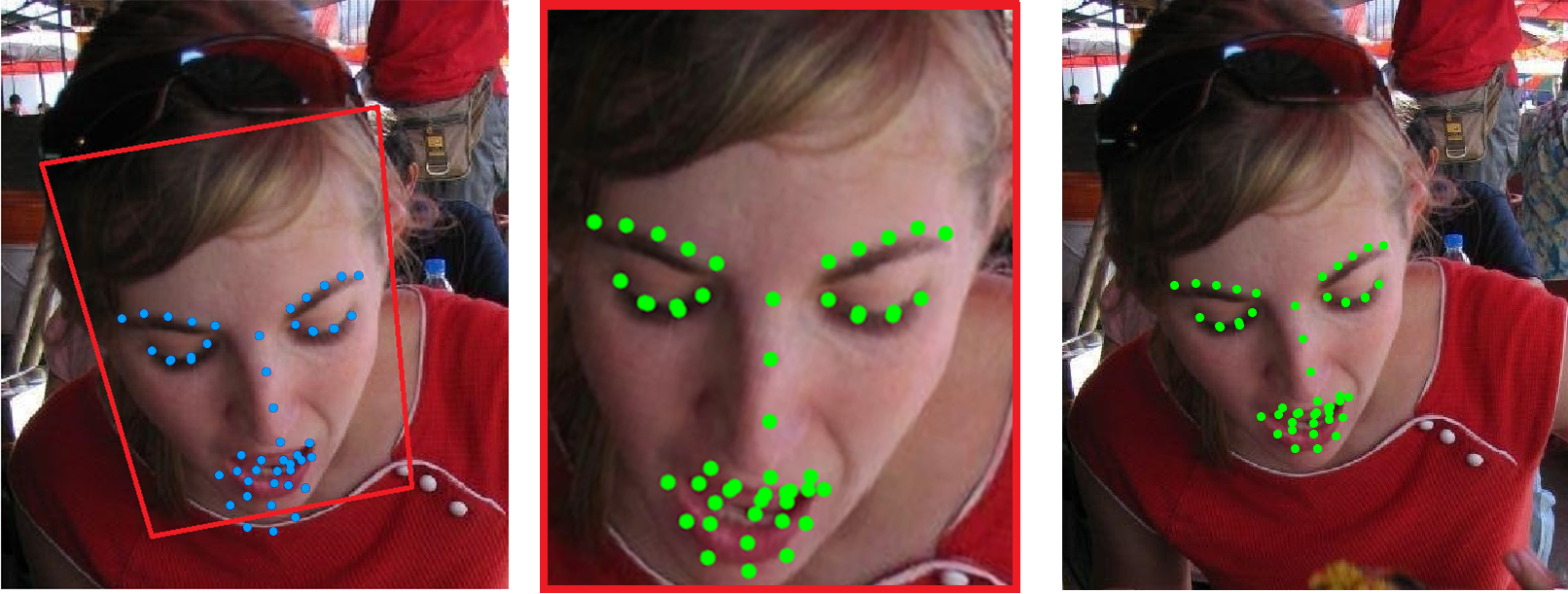}
\end{center}
   \caption{Improving the accuracy of SDM using face alignment.}
\label{fig07}
\end{figure}

For more robust modern approaches such as the Supervised Descend Method (SDM)~\cite{xiong2013supervised}, the proposed method can also be applied to improve their performance on difficult cases.
Figure~\ref{fig07} shows how the face alignment helps SDM on a face image containing an extremely large pose variation.
The accuracies of landmarks around the mouth are significantly improved.

\section{Conclusions}

We propose an end-to-end trainable framework in which the face alignment and the facial feature extraction can be jointly trained using only the personal identities as the supervising signal.
As such, explicit knowledge about human face characteristics and artificially defined geometric transformation principles are no longer needed for face alignment in the recognition task.
Our proposal actually lay a foundation for the future implementation of a fully end-to-end face recognition system which can actually be easily extended to other find grained object recognition tasks.
Another future work is to improve the robustness of the transformation prediction toward extreme pose variations and exaggerated facial expressions using more training data and more carefully designed data augmentation strategy.

{\small
\bibliographystyle{ieee}
\bibliography{egbib}
}

\end{document}